# Perturbative GAN: GAN with Perturbation Layers

Yuma Kishi[†‡] Tsutomu Ikegami[†] Shin-ichi O'uchi[†] Ryousei Takano[†] Wakana Nogami[†‡] Tomohiro Kudoh[‡†]

**Abstract**

*Perturbative GAN, which replaces convolution layers of existing convolutional GANs (DCGAN, WGAN-GP, BIG-GAN, etc.) with perturbation layers that adds a fixed noise mask, is proposed. Compared with the convolutional GANs, the number of parameters to be trained is smaller, the convergence of training is faster, the inception score of generated images is higher, and the overall training cost is reduced. Algorithmic generation of the noise masks is also proposed, with which the training, as well as the generation, can be boosted with hardware acceleration. Perturbative GAN is evaluated using conventional datasets (CIFAR10, LSUN, ImageNet), both in the cases when a perturbation layer is adopted only for Generators and when it is introduced to both Generator and Discriminator.*

## I. INTRODUCTION

Generative Adversary Networks (GAN)[1] is a generative model proposed by Goodfellow et al in 2014: the Generator generates fake data having the same distribution as the training data from the random noise input and the Discriminator identifies whether the input is data from the teacher data or that generated by the Generator. Based on Deep Convolutional GAN (DCGAN)[2] by Alec Radford et al in 2015 which introduced the convolution layer to the GAN field, *CNN-based* GAN models have shown remarkable results in image generation, style transformation, and speech generation. Recently, it has been found that high definition images of $1024 \times 1024$ can be generated without making the stack structure deeper.

Training a GAN is very unstable compared to other generation models. With Generator $G$ and Discriminator $D$, the learning process of a GAN can be expressed as the following minimax game [1]:

$$\min_{G} \max_{D} V(D,G) = \mathbb{E}_{x \sim P_{data}}[\log(D(x))] + \mathbb{E}_{z \sim P_z}[1 - \log(D(G(z)))] \quad (1)$$

[†]National Institute of Advanced Industrial Science and Technology, Tsukuba, Japan
[‡]The University of Tokyo, Tokyo, Japan

where $\mathbb{E}$ denotes expectation, $x$ denotes samples chosen from the real data distribution $P_{data}$, and $z$ denotes samples chosen from the fake model distribution $P_z$ implicitly defined by the Generator.

Thus, training a GAN is a task of finding a Nash equilibrium in a minimax game. However, if the Discriminator's discrimination performance overwhelms the Generator, it often leads to *vanishing gradients* for updating parameters. On the other hand, if the performance of the Discriminator is significantly inferior to that of the Generator, the Generator falls into *mode collapse*, a state where the Generator generates only a uniform image.

Training tends to become more unstable when high resolution images are generated because the larger the input image is, the easier it is for the Discriminator to discriminate between real or fake. In order to solve this problem, Tao Karras et al proposed Progressive Growing of GANs for Improved Quality, Stability, and Variation (PGGAN) [3]: The Generator and Discriminator both grow from low to high resolution as new layers are added to that model. Experimental results showed that PGGAN can generate $1024 \times 1024$ images. Andrew Brock et al showed in Large Scale GAN Training for High Fidelity Natural Image Synthesis (BIGGAN) [4] that a Generator can generate $1024 \times 1024$ images by enlarging the model size and batch size even without a stack structure. However, compared to DCGAN, the batch size is ~256 to 2048, and the number of parameters is increased from ~12 M to ~173 M. Large-scale training requires a lot of GPU resources, and the cost of learning is very large. In order to broaden the application range of GANs, demand for methods that can efficiently train with a smaller number of parameters is increasing.

On the other hand, much research on training cost reduction has been done in the classification model field. Matthieu Courbariaux et al showed in Binarized Neural Networks: Training Deep Neural Networks with Weights and Activations Constrained to +1 or -1[5] that training can be achieved for small datasets (MNIST, CIFAR10) with accuracy deterioration of about 1%, even if the parameters and activation of the conventional Convolutional Neural Networks (CNN) model are quantized to +1 or -1. In response to this research, Shuchang Zhou et al evaluated AlexNet's parameters, activation, and gradient for a large dataset (ImageNet) to reduce the bit size to several bits without lowering the accuracy of classification in DoReFa-Net: Training Low Bitwidth Convolutional Neural Networks with Low Bitwidth Gradients. [6] Among all these efforts, Felix Juefei-Xu et al proposed Perturbative Neural Networks (PNN) [7] in 2018 and showed that the number of parameters to be trained can be drastically reduced by replacing the convolution layers of CNN by perturbation layers, without sacrificing the accuracy. In the perturbation layer, fixed noise masks are applied to activations, and linear combinations are taken among them. Since only the weight of the linear combination is trained, the number of parameters to be trained is reduced by a square of the kernel size as compared with CNN. Furthermore, since convolution operations are replaced by the simple addition of noise, processing can be performed faster than with the conventional method, especially when implementation in hardware, such as an FPGA, is considered.

In this paper, we propose a **Perturbative GAN** (**PGAN**) in which the convolution layers of the GAN are replaced by the perturbation layers. The performance of the Perturbative GAN was assessed with the conventional datasets of Cifar10, LSUN, and ImageNet, for two cases where the perturbation layers are introduced only to the Generator, or to both the Generator and Discriminator. In either case, the Inception Score of the generated images is improved from the conventional GAN, while convergence of the training becomes faster. In addition, algorithmic generation of the fixed noise masks is proposed, with which bulky noise masks are not stored in memory but generated on-the-fly.

The main contributions of this paper are as follows:

1. We propose a Perturbative GAN (PGAN) with a perturbation layer instead of a convolution layer.
2. Improvement of the Inception Score and reduction of the training cost are demonstrated by assessing PGAN with the conventional datasets (cifar10, LSUN, ImageNet)
3. Algorithmic generation of noise masks is proposed, aiming at the hardware acceleration of PGAN.

The rest of the paper is organized as follows;
In Chapter 2, we review the background researches on this method. In Chapter 3, we describe our proposed method, PGAN, where perturbation layers are introduced in a GAN. In Chapter 4, the details of PGAN implementation and evaluation are shown, followed by the discussion on the results in Chapter 5. Finally, Chapter 6 concludes the paper.

## II. BACKGROUND

Since PGAN is mainly based on the idea of PNN and LBCNN, this chapter describes the main concept of PNN and LBCNN.

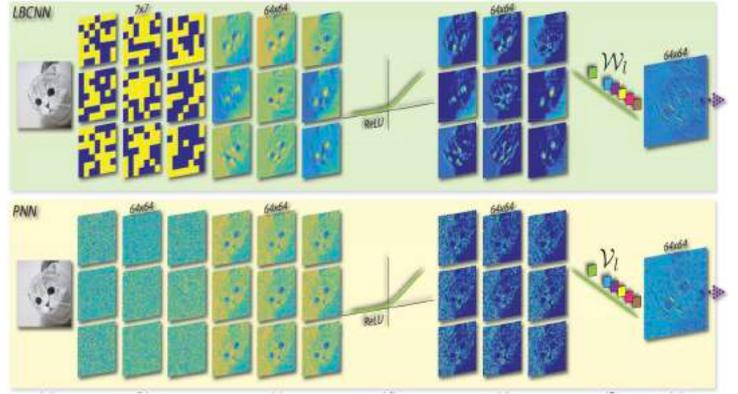

**Figure 1:** Basic modules in LBCNN [8], and PNN [7]. $\mathcal{W}$ and $\mathcal{V}$ are the trainable parameters for the local binary convolution layer and the perturbation layer. (This image is from [7])

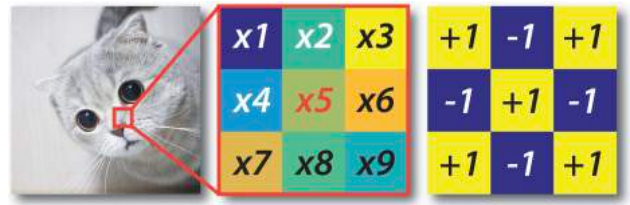

**Figure 2:** Local binary convolution in LBCNN is equivalent to addition and subtraction among eight neighborhoods. (This image is from [7])

PNN is based on the Local Binary Convolutional Neural Networks (LBCNN) concept [8], in which the feature map of the input image is calculated by using local binary patterns (LBP) [21]. The basic LBCNN module has fixed randomly generated sparse binary filters and trainable linear combination weights. The basic processing of LBCNN is performed as shown in Fig.1 (upper). First, the input image is convolved with LBP filters and the result is propagated with an activation function like ReLU, and then the activated map is linearly combined to generate the output to the next layer. These linearly produced combination weights are the only trainable parameters.

This process can be represented by the following equation [7];

$$x_{l+1}^t = \sum_{i=0}^{m} \sigma_{relu}\left(\sum_s b_{l,i}^s * x_l^s\right) \cdot \mathcal{W}_{l,i}^t \qquad (2)$$

where $t$ is the output channel, $s$ is the input channel, and $*$ is the channel-wise convolution operation. Again, linear weights $\mathcal{W}$ are the only trainable parameters of an LBCNN layer. As can be seen from equation (2), the number of parameters to be trained is smaller than that of the conventional CNN. Furthermore, LBCNN has also succeeded in simplifying the complicated convolution process to the addition and subtraction of adjacent pixels by LBP.

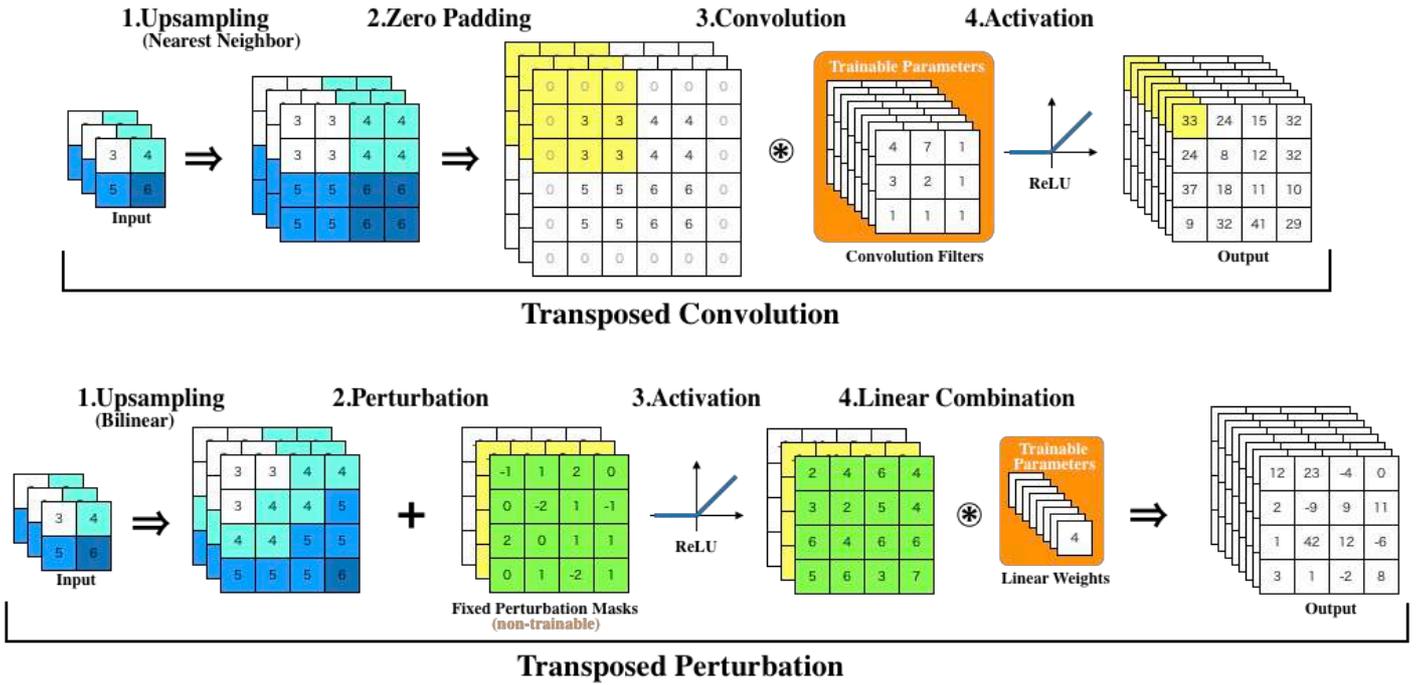

**Figure 3:** Basic modules in CNN-based Generator (upper) and PNN-based Generator (bottom). In Transposed Convolution: 1. Input is upsampled by using nearest neighbor method. 2. Zero padding is done to adjust the output size. 3. A convolution operation is performed with kernel filters having trainable parameters. 4. The convolution result is propagated to the activation function, and the activated feature map is output to the next layer. In Transposed Perturbation: 1. Input is upsampled by using bilinear method. 2. Non-trainable fixed noise masks of the same size are added to the upsampled result. 3. The addition result is propagated to the activation function. 4. The activated feature maps are linearly combined and output to the next layer. Only the weights of the linear combination at this time is trained.

The achievement of LBCNN is based on the reproduction of the same processing as the conventional convolution processing by using the fixed LBP filters and the linear combination weights. As shown in Fig.2, 3×3 pixels are extracted from the input image, named $x_1, \ldots, x_9$ now, and are multiplied by the LBP filter, components of which are either $+1$ or $-1$. The result of the convolution multiplication becomes, like, for example, $y = x_1 + x_3 + x_5 + x_7 + x_9 - x_2 - x_4 - x_6 - x_8$, and the same process repeats until the entire response map is calculated. Because these results with different LBP filters are linearly combined with weight factors, Here, this process is similar to the convolution centered on $x_5$. Since the conventional convolution processing is also the result of adding and subtracting the neighboring pixels of the center pixel, we can see that the operation of LBCNN is closer to the conventional CNN convolution processing when considering linear combination.

Based on this idea, PNN simplified the convolution process in LBCNN by the addition of fixed noise masks. The basic processing of PNN is performed as shown in Fig.3. Instead of the LBP convolution, fixed noise masks are added to the input image. The subsequent processing is the same as that of LBCNN; the perturbated result is propagated with an activation function like ReLU, and then the activated map is linearly combined to generate the output to the next layer. This process can be represented by the following equation [7];

$$x_{l+1}^t = \sum_{i=0}^{m} \sigma_{relu}(\mathcal{N}_l^i + x_l^i) \cdot \mathcal{V}_{l,i}^t \quad (3)$$

where $t$ is the output channel, $i$ is the input channel, and $\mathcal{N}_l^i$ is the $i$-th random additive perturbation mask in layer $l$. Similar to LBCNN, linear weights $\mathcal{V}$ are the only trainable parameters of a PNN layer. Interestingly, although PNN replaces the elaborated convolution process in LBCNN by the simple addition of a noise mask, PNN achieves slightly better classification accuracy than the conventional method in the ImageNet dataset.

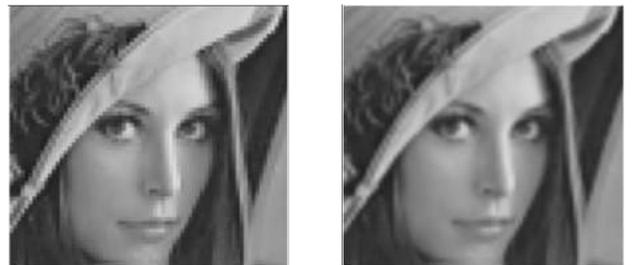

**Figure 4:** Upsampling Examples obtained by using Nearest Neighbor (Left) and Bilinear Interpolation

## III. PROPOSED METHOD

Inspired by the success of PNN in the image classification stage, we propose the introduction of the perturbation layers into the generation model of GAN.

### 3.1. Introducing Perturbation Layers to GAN

The basic Generator operation of GAN is a transpose of convolution-pooling process, which transforms a random noise to a fake image. As shown in the upper part of Fig. 3, the input noise is first unsampled, zero padded, and convolved with filters. The result is then propagated to the activation function and supplied as input to the next layer.

Therefore, in order to introduce the perturbation layer into the Generator, it suffices to replace the convolution process of the transposed convolution process with the noise addition process and linear combination, as shown in the bottom part of Fig.3. In mathematical terms, this can be expressed using equation (4);

$$x^t_{l+1} = \sum_{i=0}^{m} \sigma_{relu}\left(\mathcal{N}^i_l + u_{bilinear}(x^i_l)\right) \cdot \mathcal{V}^t_{l,i} \quad (4)$$

where $t$ is the output channel, $i$ is the input channel, and $\mathcal{N}^i_l$ is the $i$-th random additive perturbation mask in layer $l$. $u_{bilinear}$ denotes upsampling by bilinear interpolation. In the CNN-based GAN, the nearest neighbor method is often used as the upsampling technique. As a mathematical expression, the nearest neighbor method $u_{nearest}$ can be expressed as follows [9];

$$u_{nearest}(x, y) = ([x + 0.5], [y + 0.5]) \quad (5)$$

where [ ] represents floor processing. Although this method is simple to implement and fast, mosaic noise is noticeable (Fig.4 Left).

On the other hand, the bilinear method $u_{bilinear}$ can be expressed as follows [10];

$$u_{bilinear}(x, y) \approx \frac{y_2 - y}{y_2 - y_1} f(x, y_1) + \frac{y - y_1}{y_2 - y_1} f(x, y_2)$$
$$= \frac{y_2 - y}{y_2 - y_1}\left(\frac{x_2 - x}{x_2 - x_1} f(x_1, y_1) + \frac{x - x_1}{x_2 - x_1} f(x_2, y_1)\right)$$
$$+ \frac{y - y_1}{y_2 - y_1}\left(\frac{x_2 - x}{x_2 - x_1} f(x_1, y_2) + \frac{x - x_1}{x_2 - x_1} f(x_2, y_2)\right) \quad (6)$$

where $f(x_1, y_1), f(x_1, y_2), f(x_2, y_1), f(x_2, y_2)$ denote the four neighborhoods of the target pixel. As can be seen from the comparison of equations (5) and (6), bilinear interpolation is computationally more demanding than nearest neighbor interpolation, but the resulting upsampled images are smoother. (Fig.4 Right)

In PGAN, we noticed that the mosaic noise causes instability of learning in the layer close to the input. The adoption of the bilinear interpolation is essential to obtain a stable generation result.

For the PGAN Discriminator, we examined the possibility of both the conventional CNN-based Discriminator and the introduced PNN-based Discriminator utilizing the perturbation layer. The PNN-based Discriminator adopts a slightly modified version of an existing implementation of PNN [7].

### 3.2. Algorithmic Generation of Noise

In this section we will discuss how to generate fixed noise masks. In PNN's official PyTorch implementation, the fixed noise masks are generated from the standard normal distribution using the PyTorch function. However, if this can be further simplified to a uniform distribution, it can be expected to further speed up the operation when implemented in hardware such as an FPGA.

We investigated (1) the Linear Congruential (LC) Generator [11] method and (2) the Mersenne Twister (MT) [12] method as uniformly distributed noise generation methods.

(1) Linear Congruential Generator

LC is the simplest and fastest method for generating pseudorandom numbers. the Generator is defined by a recurrence relation [11]:

$$x_{n+1} = (ax_n + c) \bmod m \quad (7)$$

where $x$ denotes pseudorandom values, $a, c, m$ are constants, and pseudorandom numbers can be generated with the maximum period $2^{32}$ by choosing an appropriate number. Although LC is the easiest way, it is known that when random numbers are generated for multidimensional arrays, non-randomness occurs due to low periodicity. (Fig.5)

(2) Mersenne Twister

MT is a pseudorandom number generation method widely used now, available in various languages such as C ++, PHP, and Python. The periodicity of MT is significantly improved from the linear recurrence formula, which can reach as large as $2^{11937} - 1$ and higher. The detail of the expression is omitted here.

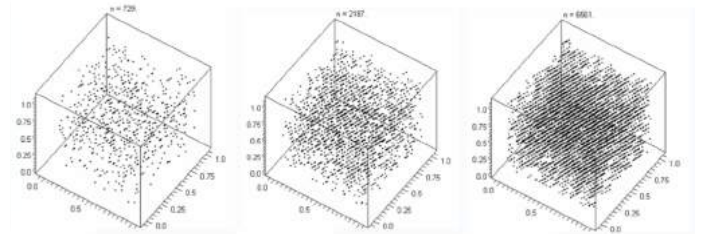

**Figure 5:** Example of Plot of Pseudo Random Number in 3D Space by LC. When the number of plots increases, a striped pattern emerges. [19]

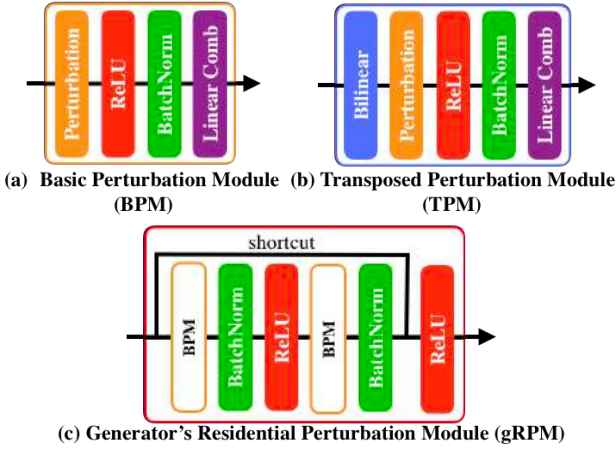

**Figure 6:** Details of (a)Basic Perturbation Module (BPM) and (b)Transposed Perturbation Module (TPM) (c) Generator's Residential Perturbation Module (gRPM)

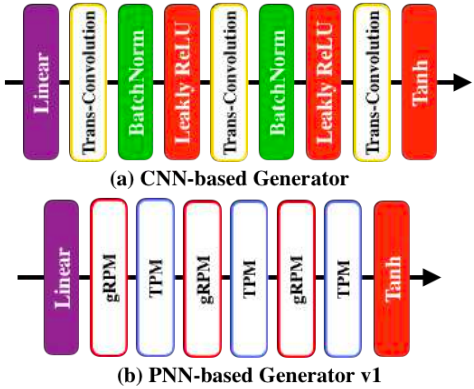

**Figure 7:** Implementation Details of (a)CNN-based Generator (CG) and (b)PNN-based Generator v1 (PGv1). In both models, latent variables of size ($batch\ size, 128$) are resized to size ($batch\ size, 128 \times 8, 4, 4$) by the first linear layer. PGv2 has one more gRPM and TPM because the first layer resizes to ($batch\ size, 128 \times 8, 2, 2$). PGv3 has two more each module because the first layer resizes to ($batch\ size, 128 \times 8, 1, 1$).

**Table 1:** The number of parameters and the ratio between each Generator models.

| model | # of params | ratio with CG |
|---|---|---|
| CG | 11,540,480 | 1.00 |
| PGv1 | 5,570,944 | 0.48 |
| PGv2 | 5,833,088 | 0.51 |
| PGv3 | 7,537,024 | 0.65 |

## IV. EXPERIMENTS

This chapter describes experiments performed to evaluate PGAN. Three experiments were conducted: (1) Training PNN-based Generator and CNN-based Discriminator, (2) Training a PNN-based Generator and PNN-based Discriminator, and (3) Noise mask generation by an algorithm. The datasets used in the experiment were cifar10 [15], LSUN [16], and ImageNet [17]. In the experiment with the LSUN dataset, we used approximately 3M images of the class of "bedroom" which is resized to $32 \times 32$. With the ImageNet dataset, 1.2 M images of 1,000 classes resized to $32 \times 32$ were used for training. The size of the generated image is $32 \times 32$, the batch size is basically fixed at 64, 128 dimensional random vectors of the standard normal distribution are taken as input latent variables. Basic algorithms and loss functions of training were adopted from WGAN-gp [13]. The Inception Score [14] is employed as the evaluation index, where the number of generated images is $N = 1024$.

### 4.1. PNN-based Generator and CNN-based Discriminator

PNN-based Generator (PG) consists of a basic noise module, a residential noise block with a shortcut, and a linear layer in the first layer. The structure of basic perturbation module (BPM) and the transposed perturbation module (TPM), and the Generator's residential perturbation module (gRPM) have the structure shown in Fig 6.

Three types of PGs are examined, in which the number of parameters of the first linear layer was varied (PGv1 PGv2 PGv3). PGv1 has the largest number of parameters of the linear layer, PGv3 has the smallest. The number of parameters to be trained in PG can be significantly reduced from that in the CNN-based Generators (CG). Reduction ratio for a single layer is

$$\frac{\#param\ in\ PG}{\#param\ in\ CG} = \frac{p \times q}{p \times q \times k \times k} = \frac{1}{k^2} \qquad (8)$$

where $p$ and $q$ are the number of input and output channels, respectively, and $k$ is the kernel size of the convolution layer. Note that, although the reduction ratio shown above is achieved for simple datasets like MNIST, the actual reduction ratio in the image generation is less drastic, because larger $p$ and $q$ are typically required for PG.

The composition of layers is shown in Fig.7 and the number of parameters is shown in Table 1 for PG and CG.

**Table 2:** Inception Score for Evaluation of PNN-based Generator

| dataset | PGv1 vs. CD | PGv2 vs. CD | PGv3 vs. CD | CG vs. CD |
|---|---|---|---|---|
| cifar10 | 6.0 (±0.5) | 5.3 (±0.5) | 4.3 (±0.3) | 5.3 (±0.2) |
| LSUN | 3.8 (±0.2) | 3.8 (±0.2) | 3.3 (±0.5) | 3.2 (±0.2) |
| ImageNet | 6.5 (±0.5) | 6.5 (±0.3) | 4.2 (±0.5) | 4.5 (±0.3) |

For the CNN-based Discriminator (CD) uses the WGAN-gp[13] model composed from a linear layer and three convolution layers.

These models are trained by using cifar10, LSUN, and ImageNet datasets, individually, and are evaluated based on the Inception Score. The results are summarized in Table 2. In most cases, PG models acquire better Inception Scores than the conventional CG model. Especially, Inception Score of PGv1 is improved by 57% from CG for the ImageNet dataset.

Learning curves of these models are also shown for cifar10 and ImageNet datasets in Fig.10(a) and (d), respectively, where Inception Score is plotted for each epoch (or iteration). Obviously, PG converges faster than CG, especially for the ImageNet dataset. Note that hyper parameters are not tuned yet for the PG trainings.

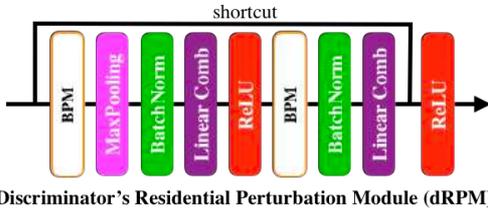

**Figure 8:** Details of the Discriminator's Residential Perturbation Module (dRPM)

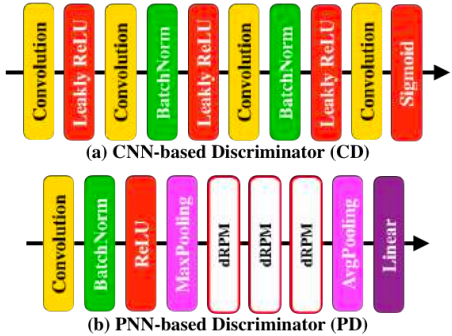

**Figure 9:** Implementation Details of (a) the CNN-based Discriminator (CG) and (b) the PNN-based Discriminator.

**Table 3:** The number of parameters and the ratio between each Generator models.

| model | # of params | ratio with CG |
|---|---|---|
| CD | 11,017,216 | 1.00 |
| PD | 1,395,584 | 0.13 |

**4.2. PNN-based Generator and PNN-based Discriminator**

Next, we introduce the perturbation layers to the Discriminator. Implementation of the PNN-based Discriminator (PD) is taken from the PyTorch implementation based on ResNet [20] by Felix Juefei-Xu, which is modified for the input image size of $32 \times 32$. The basic module of PD is represented in Fig.8. For the Generator side, we use PGv1 because it marks the highest score in the previous section.

The model is trained and evaluated in the same way as in the previous section. The results are summarized in Table 4. When PD is adopted, the Inception Score decreases slightly from the PG vs CD case but is still higher than the conventional method (CG vs CD): Inception Score is improved by 50% from the conventional method. Fast convergence of PG is also existent, as shown in Fig.10(b) and (e)

**Table 4:** Inception Score for Evaluation of PNN-based Discriminator

| dataset | PGv1 vs. PD | PGv1 vs. CD | CG vs. CD |
|---|---|---|---|
| cifar10 | 5.6 (±0.3) | 6.0 (±0.5) | 5.3 (±0.2) |
| LSUN | 3.6 (±0.2) | 3.8 (±0.2) | 3.2 (±0.2) |
| ImageNet | 6.5 (±0.5) | 6.5 (±0.5) | 4.5 (±0.3) |

Looking at the transition of the Inception Score, it was found that the adoption of PD also showed a higher increase in the Inception Score than that of the conventional method. The generated images are compared between PG and CG for several training stages in Fig.11.

**4.3. Algorithmic Generation of Noise Masks**

In the experiments shown in Sections 4.1 and 4.2, the noise masks are generated by the standard PyTorch facility and stored as non-trainable parameters. A random number Generator (RNG) of the standard normal distribution (SND) is employed there, which is backed up by the Mersenne Twister RNG. In this section, we assess the algorithmic generation of the noise masks, where the nose masks are generated on-the-fly by using the less elaborated RNG. The PGv1 vs PD model is used for the evaluation, where noise masks of both Generator and Discriminator are generated by algorithm. Two RNGs of the uniform distribution (UD) are examined: the Mersenne Twister (MT) RNG and the Linear Congruential (LC) RNG. The results are summarized in Table 5.

**Table 5:** Inception Score for Evaluation of How to Generate Noise Masks

| dataset | MT(SND) | MT(UD) | LC(UD) |
|---|---|---|---|
| cifar10 | 5.6 (±0.3) | 5.6 (±0.3) | 4.5 (±1.0) |
| LSUN | 3.6 (±0.2) | 3.8 (±0.1) | 3.2 (±0.4) |
| ImageNet | 6.5 (±0.5) | 5.1 (±0.9) | 6.0 (±0.3) |

It can be seen that Inception Score is indifferent to the distribution of the random number, but is sensitive to the quality of the random number. When LC-RNG is used to generate noise masks, Inception Score drops below the conventional method. Convergence of PG is also unstable for LC-RNG, as shown in Fig.10(c) and (f). It seems that the learning with the LC-RNG noise masks occasionally falls into local solutions. These degradations may be due to the short periodicity of the LC-RNG, which causes structures in noise masks as shown in Fig.5. It is our future work to investigate several countermeasures for using LC-RNG.

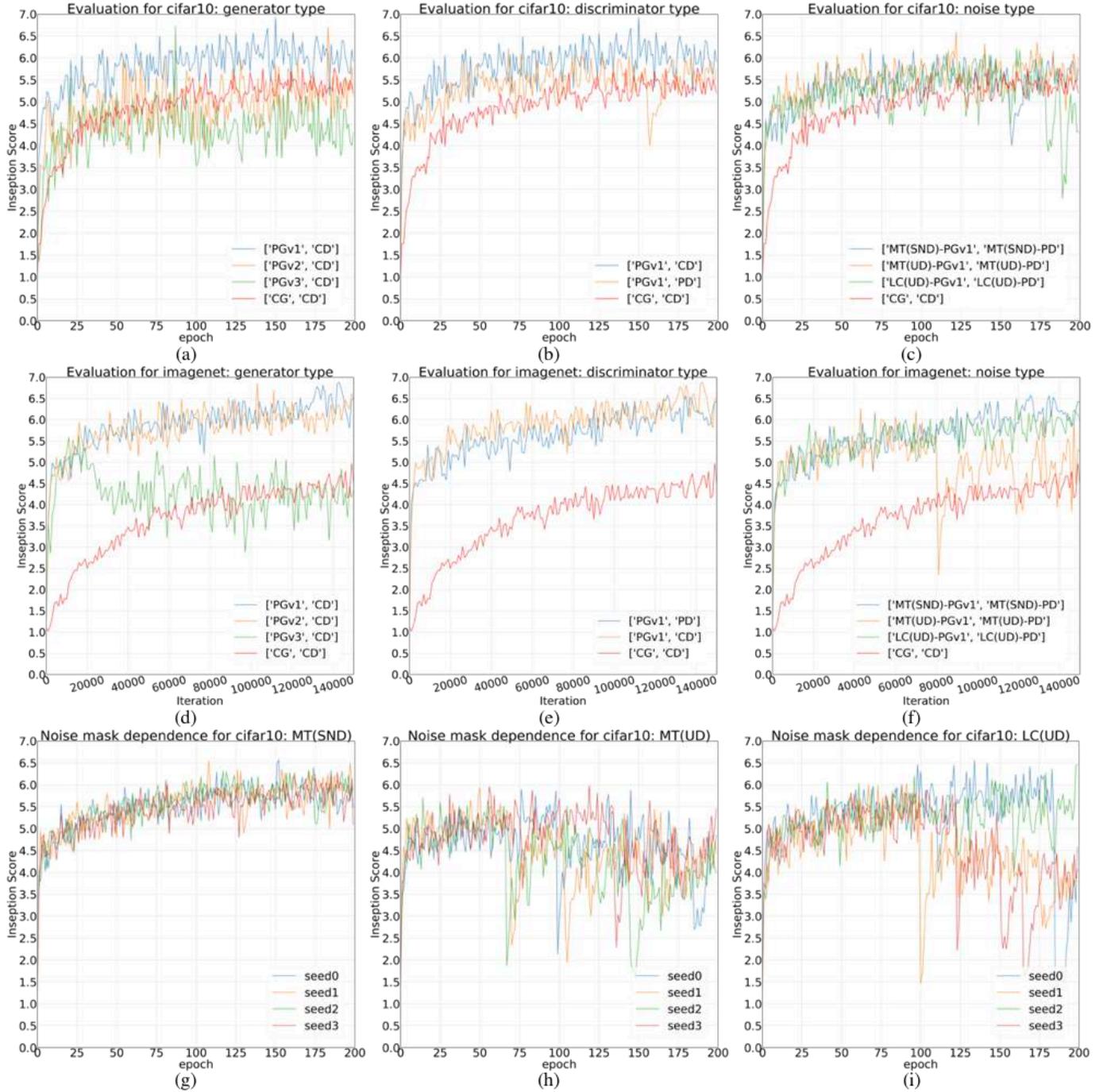

**Figure 10:** Results of evaluation experiment using Inception Score: Comparison of the Generator for (a)cifar10 and (d)ImageNet. Comparison of the Discriminator for (b)cifar10 and (e)ImageNet. Comparison of generation methods of noise for (c)cifar10 and (f)ImageNet. Noise mask dependence for cifar10 with (g)MT(SND), (h)MT(UD) and (i)LC(UD).

### 4.4. Noise mask dependence

Finally, we investigate the sensitivity of the PGAN performance on the choice of the noise masks. We take a seed to generate fixed noise masks as a hyper parameter and conduct a series of training by using the cifar10 dataset. It was found that the PGAN performance is not affected by the noise masks if they are generated by MT(SND), whereas the choice of the random seed is important if we employ UD-RNG. Interestingly, under the proper choice of the random seed, LC(UD) gives a comparable performance to MT(SND). The survey of the minimal requirement for RNG is our future work.

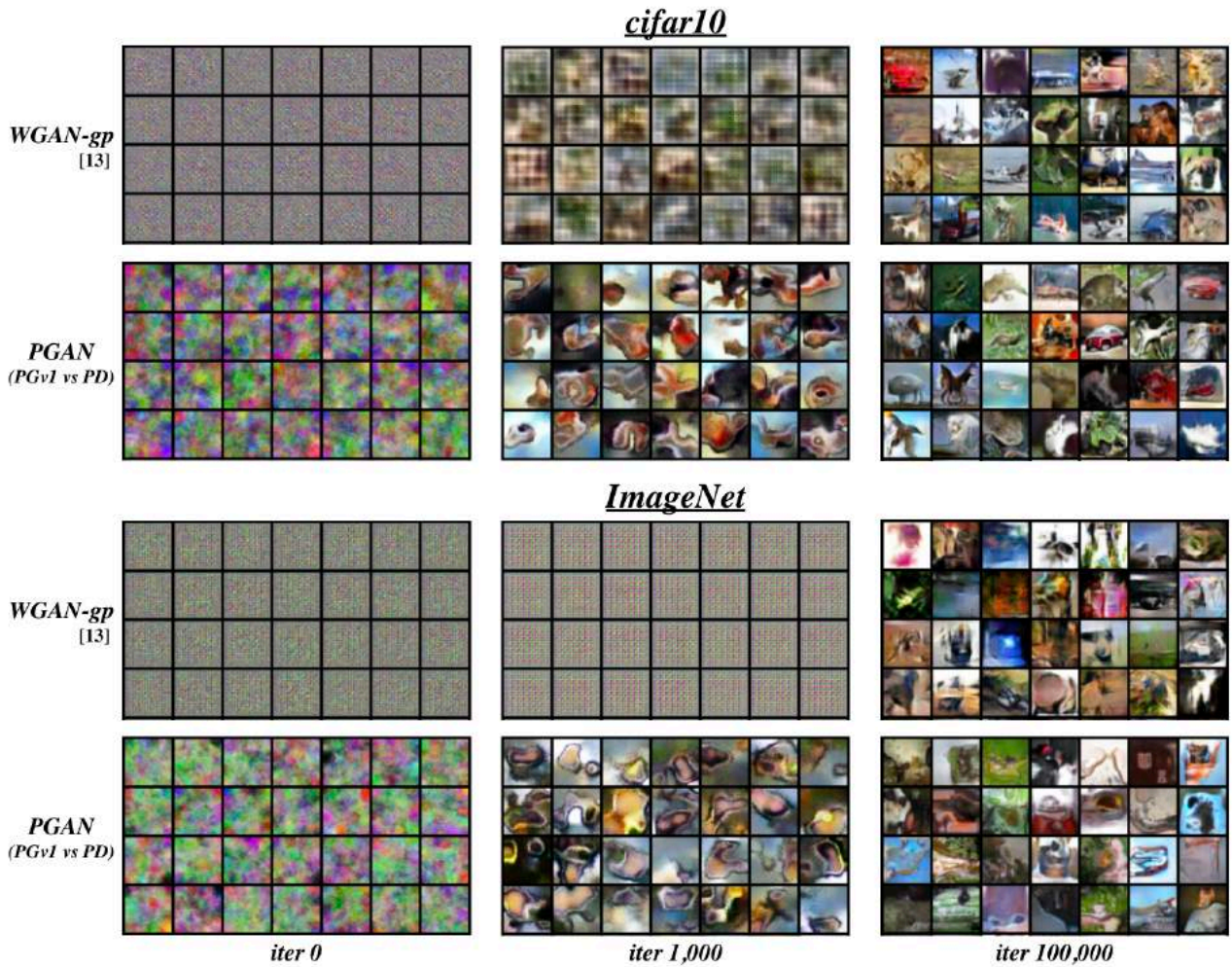

**Figure 11:** Comparison of generated fake images per iteration. Our PGAN can express the features of the training dataset at an earlier iteration than the conventional method, WGAN-gp. Looking at the generation results at the stage where learning has progressed sufficiently, it is shown that PGAN generates clear images with less blurring

## V. CONCLUTIONS

In this research, we propose Perturbative GAN (PGAN) that uses the perturbation layer in place of the convolution layer. The transposed perturbation layer is the basic module of PGAN Generator, which upsamples the input features using bilinear interpolation and adds a fixed noise mask of the same size. The resulting noisy features are propagated to the activation function, which are linearly combined and are served as input to the next layer. The weight factors of the linear combination are the only trainable parameters in the Generator of PGAN, whereas kernels of the transposed convolution layer have to be trained in the conventional GAN. Evaluation against the conventional method showed that the number of trainable parameters can be reduced to half of the CNN-based GAN and PGAN can mark a higher Inception Score for each dataset (cifar10, LSUN, ImageNet) in both cases where the perturbation layer was introduced only to the Generator and when it was introduced to both the Generator and the Discriminator. In order to save memories to store the fixed noise masks, we also investigated an algorithmic generation of the noise masks. As a result, we noticed that the quality of the random number is important in PGAN, revealing that the linear congruential Generator is not adequate. In the future we will investigate sufficient algorithm for the noise mask generation, aiming at hardware acceleration of PGAN.

## ACKNOWLEDGEMENT

This paper is based on results obtained from a project by the New Energy and Industrial Technology Development Organization (NEDO).